\documentclass{article}
\usepackage{spconf,amsmath,graphicx,glossaries}
\usepackage{amssymb}
\usepackage{mathtools}
\usepackage{tikz}
\usepackage{pgfplots}

\usepackage{orcidlink}
\usepackage{subcaption}
\usetikzlibrary{pgfplots.groupplots}
\newacronym{gan}{GAN}{Generative Adversarial Network}
\newacronym{mse}{MSE}{Mean Squared Error}
\newacronym{dm}{DM}{Diffusion Model}
\newacronym{mc}{MC}{Markov Chain}
\newacronym{vlb}{VLB}{Variational Lower Bound}
\newacronym{lpips}{LPIPS}{Learned Perceptual Image Path Similarity}
\newacronym{rdp}{RDP}{Rate-Distortion-Perception}
\newacronym{rd}{RD}{Rate-Distortion}
\newacronym{dp}{DP}{Distortion-Perception}
\newacronym{msh}{MSH}{Mean and Scale Hyperprior}

\title{Enhancing the Rate-Distortion-Perception Flexibility of Learned Image Codecs with Conditional Diffusion Decoders}
\name{Daniele Mari \thanks{Daniele Mari's activities were supported by Fondazione CaRiPaRo under the grants “Dottorati di Ricerca” 2021/2022.}\thanks{The source code for this paper can be found at https://github.com/Dan8991/Image-coding-perceptual-enhancement-with-diffusion-models.} \orcidlink{0000-0003-0727-3725}, Simone Milani \orcidlink{0000-0001-8266-5839}}

\address{University of Padova, \{daniele.mari, simone.milani\}@dei.unipd.it}

\begin{document}

\maketitle

\begin{abstract}
Learned image compression codecs have recently achieved impressive compression performances surpassing the most efficient image coding architectures. However, most approaches are trained to minimize rate and distortion which often leads to unsatisfactory visual results at low bitrates since perceptual metrics are not taken into account.
In this paper, we show that conditional diffusion models can lead to promising results in the generative compression task when used as a decoder, and that, given a compressed representation, they allow creating new tradeoff points between distortion and perception at the decoder side based on the sampling method.

\end{abstract}


\section{Introduction}
\label{sec:intro}
Since the very beginning of multimedia communications, the transmission and storage of contents like images, videos, and 3D models have always been consuming a lot of bandwidth and memory space. For these reasons, different coding architectures have been designed and deployed during the last 30 years starting from the earliest transform coding paradigms \cite{wallace1992jpeg} up to the most recent adaptive prediction methods.

In recent years, the research on image compression has gradually shifted towards the usage of neural networks \cite{balle2018variational, minnen2018joint} outperforming even the most efficient codecs, such as VVC \cite{cheng2020learned}.

Unfortunately, at very low bit rates these models show significant limitations since the decoded images report highly visible blocking artifacts in the case of non-learned methods, or blurred regions, whenever using deep learning techniques. To mitigate this issue a new research branch has started to focus on the triplet \gls{rdp} tradeoff. Perception is defined in the framework introduced by Blau et.al. \cite{blau2019rethinking} as the similarity between the distributions of the real and processed images. For this reason, while usually distortion is measured in terms of fully referenced metrics (e.g., error functions that compare two samples and tell how close they are like in \gls{mse}), perceptual quality can be measured in terms of unreferenced metrics as they aim at approximating human perceptions and reactions. Authors in  \cite{blau2019rethinking} also highlight that there is a clear trade-off between bitrate, distortion, and perceptual degradation as enhancing one of them implies reducing at least one of the others. Even though other definitions of perception can be found in the literature, in this paper we adopt the one proposed in \cite{blau2019rethinking} to allow a fair comparison with previous state-of-the-art works.

Following this idea, many works have been proposed \cite{agustsson2019generative, mentzer2020high, theis2022lossy, pan2022extreme} either based on adversarial frameworks \cite{goodfellow2020generative} or on diffusion models \cite{ho2020denoising, song2020denoising} which have recently gained a lot of attention due to the performance they displayed in image generation \cite{ho2020denoising, song2020denoising}, text-to-image synthesis \cite{nichol2021glide} and image restoration \cite{kawar2022denoising} to name a few. Such solutions also fit well in the learned image coding framework since they are modeled as variational autoencoders.

\begin{figure}
    \centering
    \includegraphics[width=.58\columnwidth]{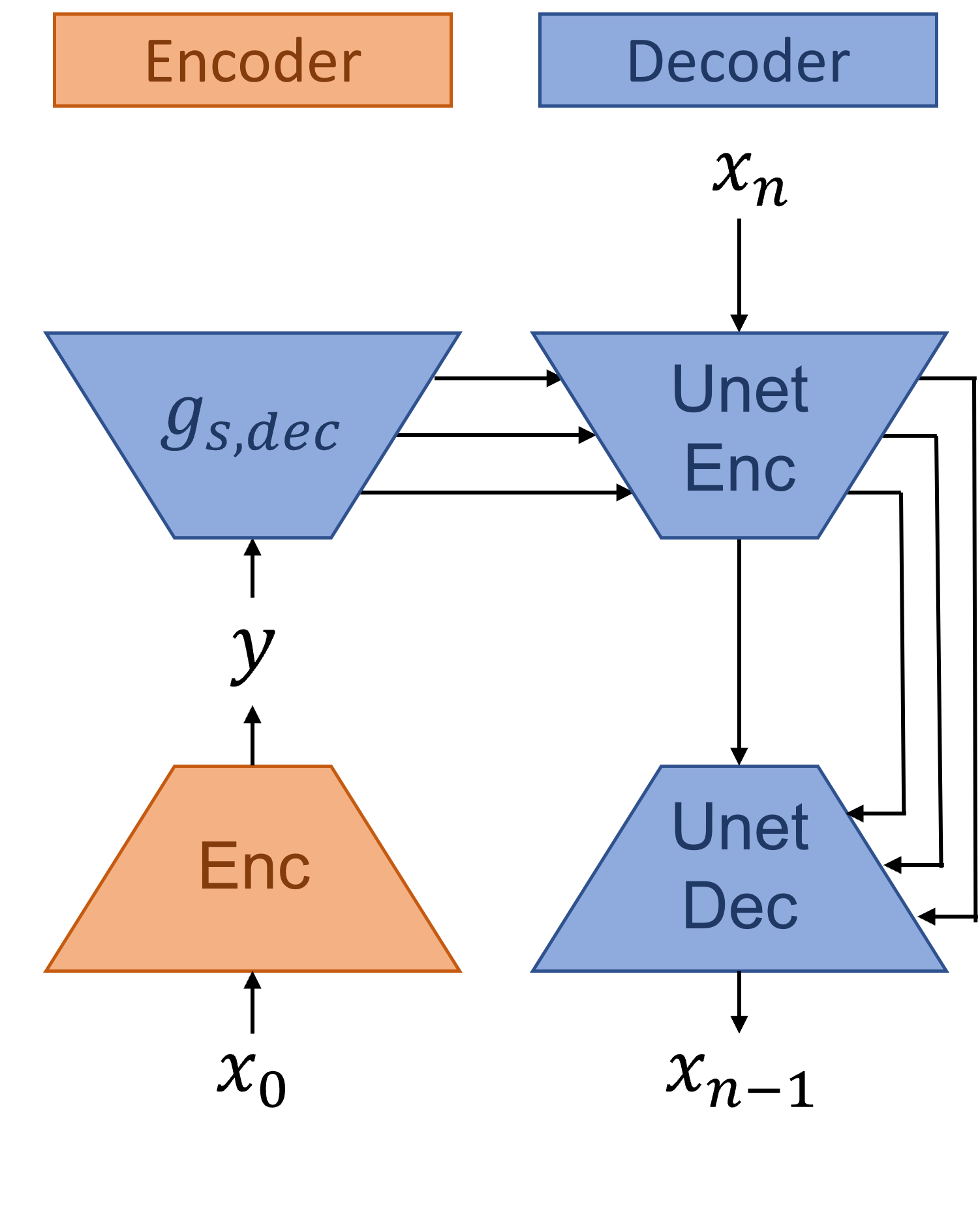}
    \vspace*{-3ex}\caption{Scheme of the proposed network}
    \label{fig:scheme}
    \vspace{-2ex}
\end{figure}

In this paper, we present a learned image coding system where at the decoder side a standard decoder (MSH \cite{balle2018variational}) and a diffusion model share the same latent space and thus can be both used for decoding. The proposed scheme achieves very promising results in terms of objective and perceptual quality. The main advantages of the proposed architecture are:

1) The encoder is derived from existing learned image codecs, i.e., the solution can be applied to any off-the-shelf coding networks without requiring a complete end-to-end re-training. Additionally, the shared latent space allows using either the distortion decoder (low complexity and distortion) or the diffusion model (higher complexity but better perceptual quality).

2) Differently from GAN-based solutions (in particular \cite{mentzer2020high}), the computational effort of reconstructions can be modulated depending on the desired quality/hardware resources by changing the diffusion sampling procedure.

3) Differently from \cite{yang2022lossy}, we show that diffusion models can produce new \gls{dp} tradeoffs by tuning the sampling method. Additionally, the shared latent space allows decoding the image with minimal distortion, if needed. 

4) On average, reconstructed images present both a high level of fidelity (with respect to the input data) and a good perceptual quality.

5) Finally we show that even if latent space was optimized using \gls{rd} metrics, perceptual quality can improve as well thanks to a suitable decoder (as envisioned in \cite{blau2019rethinking}).

\section{Related Works}
\label{sec:related}
In recent years, the most successful learned image codecs have tackled the problem of image compression using variational autoencoders \cite{balle2018variational, minnen2018joint, cheng2020learned}, where latents are regularized by means of a prior that is used to estimate the symbols probabilities. These mainly differ because of the entropy model that becomes increasingly powerful. 
Recent works have kept improving upon this by further improving the prior \cite{kim2022joint} or by tuning the features to obtain more efficient representations \cite{mari2022features}. 

Most learned approaches optimize their models using distortion metrics, however, this usually leads to unsatisfactory perceptual quality at low bitrates. For this reason, generative compression algorithms have been developed \cite{agustsson2019generative, mentzer2020high}. These typically exploit \glspl{gan} \cite{goodfellow2020generative} to reconstruct a visually pleasing image at the price of higher reconstruction error. In particular, in \cite{agustsson2019generative} the authors achieve very convincing reconstruction results at very low bitrates by training the system for \gls{rdp} with a generative loss. However, the reconstructed images usually deviate a lot from the original ones. In \cite{mentzer2020high} the authors carry out a very thorough analysis of the effects of the discriminator of the network and explore various normalization layers for regularization.

\begin{figure}
\begin{subfigure}{.49\linewidth}
  \centering
  \includegraphics[width=\linewidth]{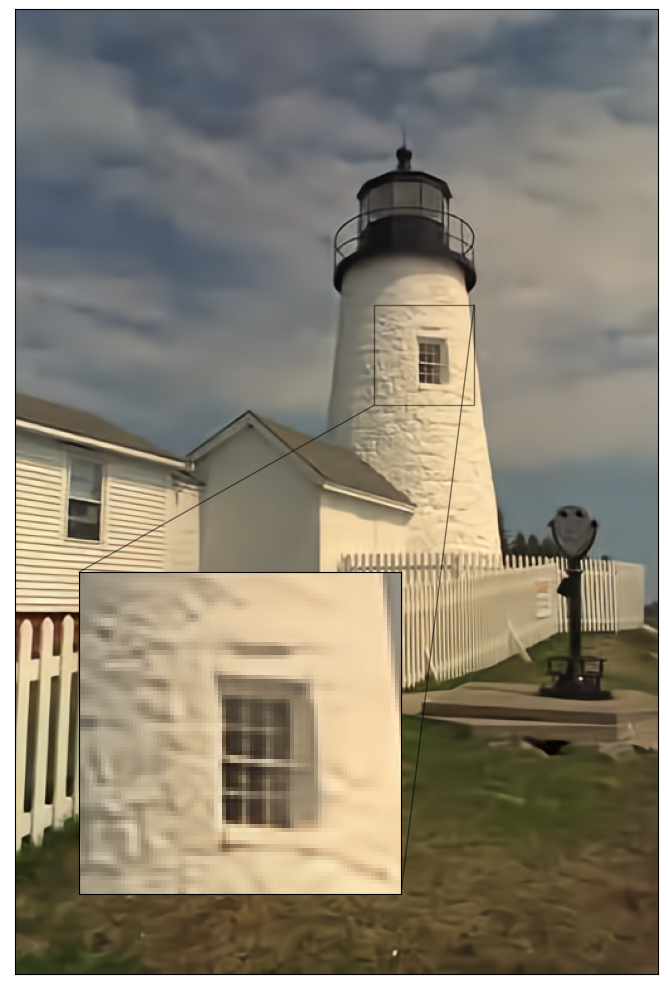}
  \caption{MSH(QP=2)}
  \label{fig:sfig1}
\end{subfigure}
\begin{subfigure}{.49\linewidth}
  \centering
  \includegraphics[width=\linewidth]{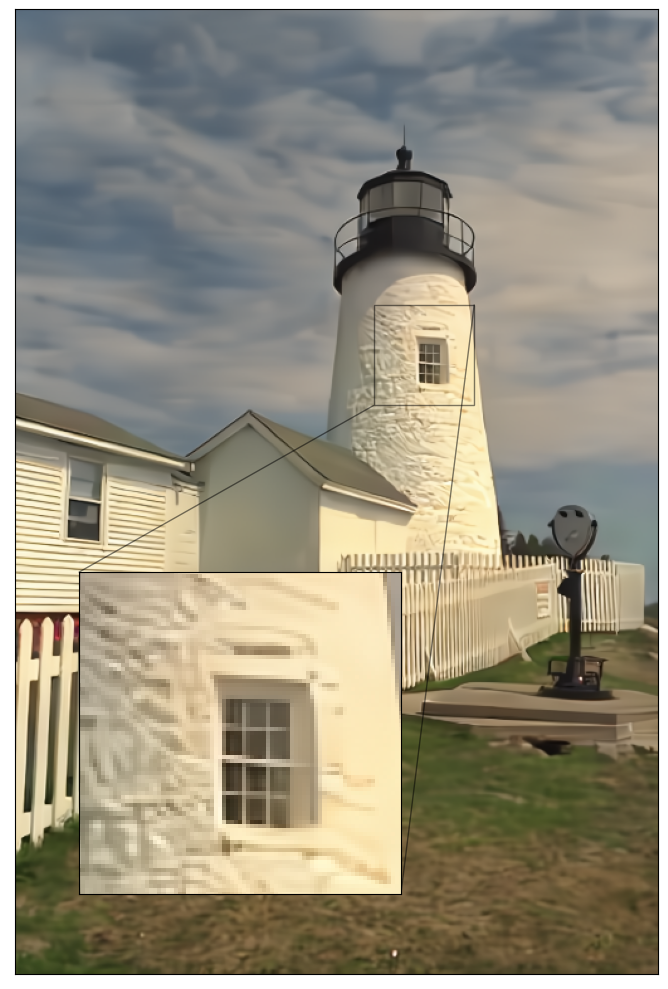}
  \caption{OURS(QP=2)}
  \label{fig:sfig2}
\end{subfigure}
\caption{Comparison between an image reconstructed with MSH fig~\ref{fig:sfig1} against the same image reconstructed with our method fig~\ref{fig:sfig2} at the exact same bitrate}
\vspace{-1.5em}
\label{fig:comparison}
\end{figure}

\glspl{dm} \cite{ho2020denoising} can also be applied to the generative compression task by conditioning them on the compressed representation of the input. The latter is used to generate an image with high likelihood that is similar to the original one.
Indeed, \glspl{dm} are well suited as decoders when using random coding \cite{li2018strong} as Theis et. al. have done in \cite{theis2022lossy}. Anyway, this approach has high computational complexity both at the encoder and at the decoder so it is better to use a standard learned encoder instead.
Among these approaches,  \cite{yang2022lossy} presents a diffusion model that is trained end-to-end from the \gls{msh} model \cite{minnen2018joint}, leading to a considerably shorter encoding time. On the other hand Pan et.al. \cite{pan2022extreme} have proposed to use a pre-trained diffusion model and to reconstruct the original image with a text description and a highly compressed version of the image as guidance. This leads to higher distortion but overall very good perceptual quality. However, in order to find the optimal textual representation text inversion is applied which is an iterative procedure that is very computationally expensive. 

\section{Method}
\label{sec:method}

Since the interest of this paper lies with conditional diffusion models, the following section will briefly present their formulation referring to \cite{ho2020denoising, song2020denoising} for a more comprehensive explanation.

Learned image codecs are usually modeled as a variational autoencoder with a learned latent prior. This allows to jointly minimize distortion of the reconstructed sample and the bitrate by optimizing for the classical loss
\begin{equation}
    \mathcal{L}(x) = \mathbb{E}_{\mathbf{x} \sim p_\mathbf{x}}\Big[d\Big(g_s(g_a(\mathbf{x})), \mathbf{x}\Big) -\lambda \log p(g_a(\mathbf{x}))\Big]
\end{equation}
where $g_a, g_s$ are the analysis and synthesis transform respectively, $d(\cdot, \cdot)$ is a distortion metric and $p(\cdot)$ is the learned prior. 

Conditional Diffusion Models \cite{ho2020denoising} are latent variables models that can be expressed as $p_\theta(\mathbf{x}_0|\mathbf{y}) = \int p_\theta(\mathbf{x}_{0:T}|\mathbf{y})d\mathbf{x}_{1:T}$ where $p_\theta(\mathbf{x}_{0:T}|\mathbf{y})$ is called the reverse process and it is modeled as a \gls{mc} with learned transition probabilities and it can be expressed as:
\begin{equation}
    p_\theta(\mathbf{x}_{0:T}|\mathbf{y}) = p(\mathbf{x}_T|\mathbf{y}) \prod_{t=1}^Tp_\theta(\mathbf{x}_{t-1}|\mathbf{x}_{t}, \mathbf{y}) 
\end{equation}
where $p(\mathbf{x}_T|\mathbf{y})=p(\mathbf{x}_T)=\mathcal{N}(\mathbf{0}, \mathbf{I})$ and
\begin{equation}
    p_\theta(\mathbf{x}_{t-1}|\mathbf{x}_{t}, \mathbf{y})=\mathcal{N}(\mathbf{\mu}_\theta(\mathbf{x}_t, \mathbf{y}, t), \mathbf{\Sigma}_\theta(\mathbf{x}_t, \mathbf{y}, t))
\end{equation}
is estimated by a neural network. The forward process is a \gls{mc} that gradually adds gaussian noise to the signal until it is completely corrupted.
The optimization of the network parameters $\theta$ can be carried out by minimizing the \gls{vlb}, however in \cite{ho2020denoising} the authors propose a more tractable approximation defined as:
\begin{equation}
    L_{simple}(\textbf{x}_0) = \mathbb{E}_{n,e}||\mathbf{\epsilon} - \mathbf{\epsilon}_\theta(\mathbf{x}_n(\mathbf{x}_0), \mathbf{y}, n)||^2
\end{equation}
where $n \sim \mathcal{U}(1,N), e\sim \mathcal{N}(0, I), \mathbf{x}_n(\mathbf{x}_0) = \sqrt{\alpha_n}\mathbf{x}_0 + \sqrt{1-\alpha_n}\mathbf{\epsilon}$ and $\alpha_n = \prod_{i=1}{n}(1-\beta_i)$ with $\beta_n \in (0,1)$ the variance schedule which can be fixed or learned. 

In this work, we use a pre-trained encoder and prior for several reasons. Firstly diffusion models are very expensive in terms of decoding time, having latents that can be decoded by a standard learned decoder gives higher flexibility to the receiver. 
The second one is that we also want to analyze from a \gls{rdp} tradeoff point of view how much the latents can affect perception. This is motivated by the fact that based on the formulation given by Blau et.al. \cite{blau2019rethinking} perception is defined as a distance between probability distributions and should thus not be affected by the quality of the conditioned latent $\mathbf{y}$.  Finally, it is easier to reach convergence without having to optimize also the rate and it removes the need to tune the $\lambda$ parameter.

The encoder and learned prior that we use are the ones from the \gls{msh} codec proposed in  \cite{minnen2018joint} and we use the models provided by Compressai \cite{begaint2020compressai}. As the synthesis transform $g_s = g_{s, unet}(\mathbf{x}_n, g_{s, dec}(\mathbf{y}))$ we use an architecture $g_{s, unet}$ based on the UNet model proposed in \cite{ho2020denoising} that is conditioned with the latents thanks to an additional decoder $g_{s, dec}$ (see Fig\ref{fig:scheme}). This proved to be more effective than simply feeding $\mathbf{y}$ in the corresponding resolution level in the UNet.
The latter has 5 resolution levels with 3 residual blocks each, we add attention modules only after the last two resolution levels. This choice was mostly taken because attention layers usually lead to very high memory utilization and we noticed that they weren't noticeably contributing to the performance of the network. As for the additional decoder, we were inspired by the solution in \cite{yang2022lossy} with 5 resolution levels each containing a Resnet block and a convolutional layer with ReLU activations. Starting from this implementation, we modified the overall structure by reducing the attention units and tuning some hyperparameters (such as the number of filters). This leads to a lower memory footprint and a simpler training process since the encoder and prior are pre-trained and fixed. This also enabled higher interoperability since the bit stream is still compatible with a standard learned decoder.

In the target function for the training of the diffusion decoder, rate does not need to be optimized, and therefore, only distortion terms are present, i.e.
\begin{equation}
    \mathcal{L}(x) = L_{lpips}(x, \mathbf{\epsilon}_\theta(\mathbf{x}_n(\mathbf{x}_0))) + L_{simple}(\textbf{x}_0)
\end{equation}
where $L_{lpips}(x, \mathbf{\epsilon}_\theta(\mathbf{x}_n(\mathbf{x}_0)))$ is the \gls{lpips} loss \cite{zhang2018unreasonable}. Note that while other works have shown that  this metric is more correlated with perceptual quality than PSNR or SSIM, it is still fully-referenced 

\section{Results}
Similarly to \cite{yang2022lossy, mentzer2020high} we only train the models at low bitrates since it is the region where perceptual quality is the most meaningful: whenever distortion becomes very small, good perceptual quality is very easy to achieve \cite{blau2019rethinking}.
For this reason, we consider the pre-trained encoders relative to quality parameters $qp \in \{1, 2, 3\}$ from the MSH model provided by the Compressai library and we keep their weights frozen during training.
The networks are trained on the ImageNet1000 dataset \cite{deng2009imagenet} for 1.5M steps by applying random cropping and rescaling to make sure that all images have the same size (128, 128) and feed the network with details at different scales. As a validation set, we use crops of the CLIC Dataset
\cite{CLIC2020} and we test on Kodak images.

\begin{figure*}[t!]
    \centering
    \includegraphics[width=\textwidth]{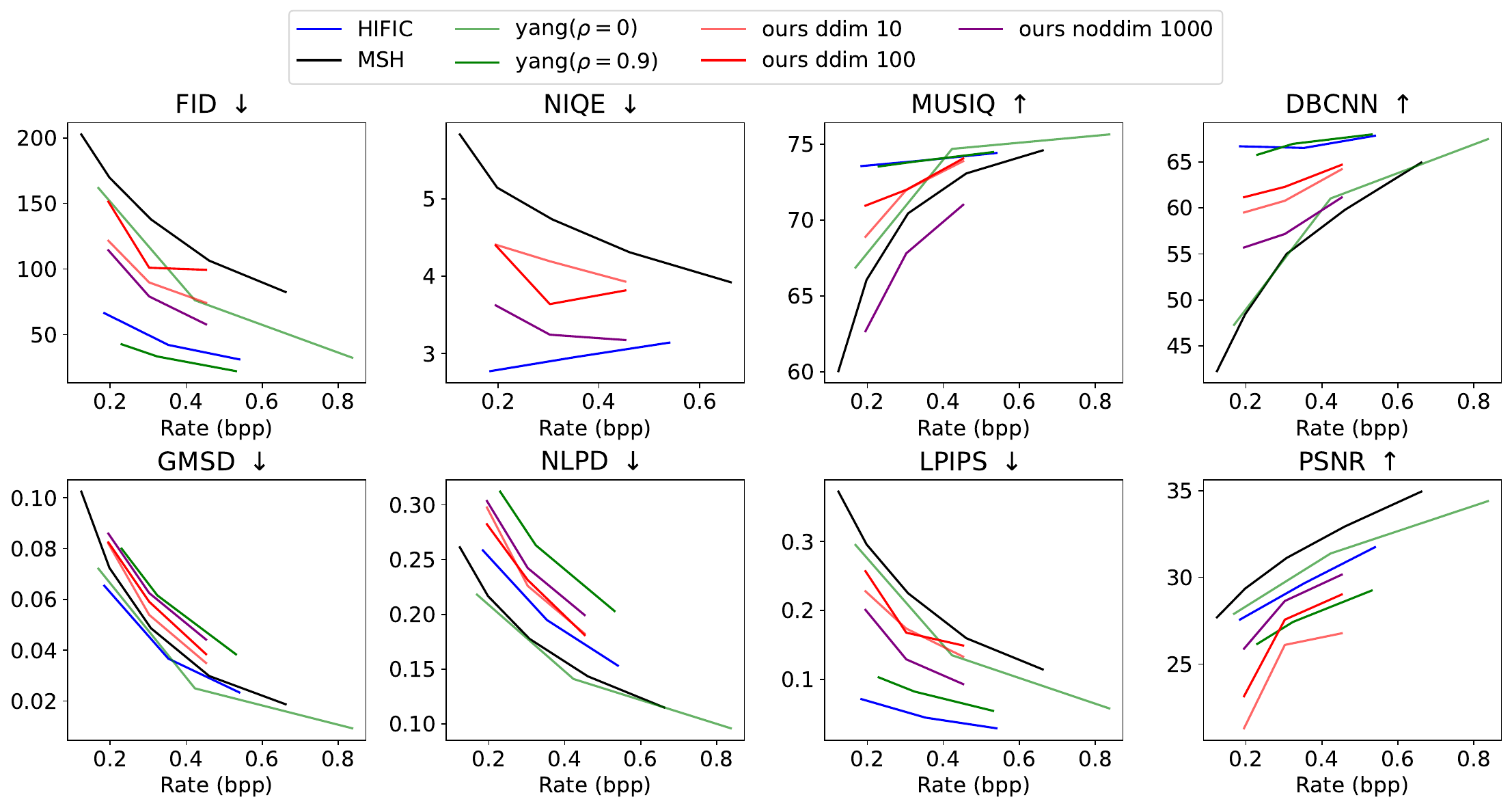}
    \caption{Comparison between results of various approaches on the Kodak dataset in terms of unreferenced and referenced metrics.}
    \label{fig:metrics}
    \vspace{-1.5em}
\end{figure*}

We compare the proposed approach against the baseline i.e. the MSH model and two state of the art generative compression algorithms i.e. HIFIC \cite{mentzer2020high} and the conditional diffusion models proposed in \cite{yang2022lossy}. 

We choose these models for comparison since they are all based on similar encoders and entropy models allowing to focus on the differences between models trained for \gls{rd} and for \gls{rdp} with both GANs and Diffusion Models.

We report results using both full-reference and no-reference metrics to capture both distortion and perception performance of the model. The referenced metrics we select are PSNR, LPIPS, NLPD, and GMSD while for unreferenced metrics we choose FID, NIQE, MUSIQ, and DBCNN. Unfortunately in \cite{yang2022lossy} results for NIQE are not reported so we omit the method in the NIQE plot (pre-trained weights are not provided by the authors, so a complete reproduction of the values in the paper is not possible). We compute FID similarly to how it was done in \cite{mentzer2020high,yang2022lossy} i.e. by splitting each image into 256x256 patches. 

In the case of \cite{yang2022lossy} results from both proposed models (trained with and without \gls{lpips}, respectively yang($\rho=0.9$), yang($\rho=0$)) are shown since the former has better perceptual quality results while the latter has better distortion results. The results relative to these plots were generated with 500 sampling steps, we refer to the original paper for additional implementation details.

We plot the metrics obtained by our models with four different sampling strategies. This is intended to show how much the sampling strategy can affect the overall performance of the diffusion model. We sampled with DDIM \cite{song2020denoising} with 10 and 100 iteration steps (more than 100 didn't really change the performance) and with DDPM with 100 and 1000 iteration steps. 

Figure \ref{fig:metrics} shows that the performance of our model lies in between the two more specialized models proposed by Yang et.al.. However, in our case, all four curves are obtained with the same 3 models simply by changing the sampling process. This ability of diffusion models to achieve different \gls{dp} tradeoffs at the same rate could be what makes them competitive with GANs for the task of generative compression. As a matter of fact, the latter have faster inference time and, by now, it shows superior performance. However, they are limited by the fact that they can produce only a single \gls{rdp} point.

Some qualitative results can be seen in figure \ref{fig:comparison}, where we compare an image compressed at the exact same rate with the MSH model (i.e. only optimized for \gls{rd}) against a sample generated by our model using DDIM and 100 steps. It is possible to see that our model produces sharper edges (see window) and more complex textures (see wall or clouds) which makes the generated sample more perceptually pleasing at the cost of higher distortion. Additionally, we noticed that the generated samples tend to have slightly different tonalities w.r.t. the original images slightly increasing distortion.

The higher performance reported by HiFIC is likely due to the adversarial training which properly optimizes the network for perceptual quality. This does not happen as much with diffusion models as proposed here and by Yang \emph{et al.} \cite{yang2022lossy} since as has been shown in the literature conditioned diffusion models tend to suffer from mode collapse and blurry artifacts (similar to the ones obtained in learned compression). Additionally according to the framework proposed in \cite{blau2019rethinking} the \gls{lpips} is not unreferenced. For this reason, we think that adding classifier (using the encoder) or classifier-free guidance might result in improved performance and in an extra degree of freedom at sampling time for extra flexibility in terms of \gls{dp} tradeoff. 

\vspace{-.5em}
\section{Conclusion}
In this work, we propose to use \gls{dm} as a decoder in the learned transform coding framework. Differently from similar approaches we keep the encoder and learned prior fixed and show that even though they were only optimized for \gls{rd} the decoder is still able to produce high perceptual quality images showing that this ability mostly depends on the decoder. This allows us to decompress the latents both using a standard learned decoder or a diffusion model which allows to choose how to trade computational resources with perceptual quality. This shows that diffusion models have great potential for image compression mostly because they allow sampling from different \gls{dp} tradeoffs by tuning the number of diffusion steps, the sampling procedure, the initialization of the diffusion latent $x_N$ (shown in \cite{yang2022lossy}), and the sampling procedure itself (e.g. DDIM, DDPM). We believe that this is the main advantage that diffusion models have w.r.t. standard codecs trained with an adversarial framework that usually need to train a different model for each \gls{rdp} tradeoff. Future works should explore the effect of the sampling process as a whole (DDIM/DDPM, number of iterations, noise initialization, etc...) and should try to introduce classifier or classifier-free guidance to improve the perceptual quality of the generated samples.

\bibliographystyle{IEEEbib}
\bibliography{biblio}

\end{document}